\documentclass[10pt,twocolumn,letterpaper]{article}

\usepackage{cvpr}
\usepackage{times}
\usepackage{epsfig}
\usepackage{graphicx}
\usepackage{subfigure}
\usepackage{amsmath}
\usepackage{amssymb}
\usepackage{multirow}
\usepackage{booktabs}
\usepackage{bbm}
\usepackage{color}
\usepackage{authblk}
\newcommand*{\affaddr}[1]{#1} 
\newcommand*{\affmark}[1][*]{\textsuperscript{#1}}

\usepackage[pagebackref=true,breaklinks=true,colorlinks,bookmarks=false]{hyperref}

\cvprfinalcopy 

\def\cvprPaperID{5747} 

\ifcvprfinal\fi
\begin{document}

\title{AdarGCN: Adaptive Aggregation GCN for Few-Shot Learning}

\author{%
Jianhong Zhang\affmark[1], Manli Zhang\affmark[1], Zhiwu Lu \thanks{Corresponding author.}\affmark[1], Tao Xiang\affmark[2], and Jirong Wen\affmark[1]\\
\affaddr{\affmark[1]Beijing Key Laboratory of Big Data Management and Analysis Methods\\
School of Information, Renmin University of China, Beijing 100872, China}\\
\affaddr{\affmark[2]Department of Electrical and Electronic Engineering, \\University of Surrey, Guildford, Surrey GU2 7XH, United Kingdom}\\
}

\maketitle

\begin{abstract}
Existing few-shot learning (FSL) methods assume that there exist sufficient training samples from source classes for knowledge transfer to target classes with few training samples. However, this assumption is often invalid, especially when it comes to fine-grained recognition. In this work, we define a new FSL setting termed few-shot few-shot learning (FSFSL), under which both the source and target classes have limited training samples. To overcome the source class data scarcity problem, a natural option is to crawl images from the web with class names as search keywords. However, the crawled images are inevitably corrupted by large amount of noise (irrelevant images) and thus may harm the performance. To address this problem, we propose a graph convolutional network (GCN)-based label denoising (LDN) method to remove the irrelevant images. Further, with the cleaned web images as well as the original clean training images, we propose a GCN-based FSL method. For both the LDN and FSL tasks, a novel adaptive aggregation GCN (AdarGCN) model is proposed, which differs from existing GCN models in that adaptive aggregation is performed based on a multi-head multi-level aggregation module. With AdarGCN, how much and how far information carried by each graph node is propagated in the graph structure can be determined automatically, therefore alleviating the effects of both noisy and outlying training samples. Extensive experiments show the superior performance of our AdarGCN under both the new FSFSL and the conventional FSL settings.
\end{abstract}

\section{Introduction}

Few-shot learning (FSL) \cite{feifei2006pami,fe2003bayesian} becomes topical. It aims to recognize a set of target classes by learning with sufficient labelled samples from a set of source classes and only few labelled samples from the target classes. Existing FSL methods \cite{Reptile,Mini_split,MAML,ProtoNet,MatchingNet,RelationNet,IMP,Peng_2019_ICCV}  employ a deep neural network (DNN) model \cite{lecun2015deep,yosinski2014transferable,he2016deep} as the backbone for FSL. They thus make the implicit assumption that there are sufficient training samples from the source classes for knowledge transfer to the target.   However, this assumption is often invalid in practice especially when it comes to fine-grained recognition. For this problem, the source classes are also fine-grained, so collecting and labeling sufficient samples for each source class is also difficult. For example, in the widely-used CUB dataset \cite{CUB-200-2011}, each bird class has less than 60 samples. Without sufficient labelled samples from source classes, it becomes harder to recognize the target classes by knowledge transfer from source classes. 

\begin{figure}[t]
\vspace{0.05in}
\centering
\includegraphics[width=0.99\columnwidth]{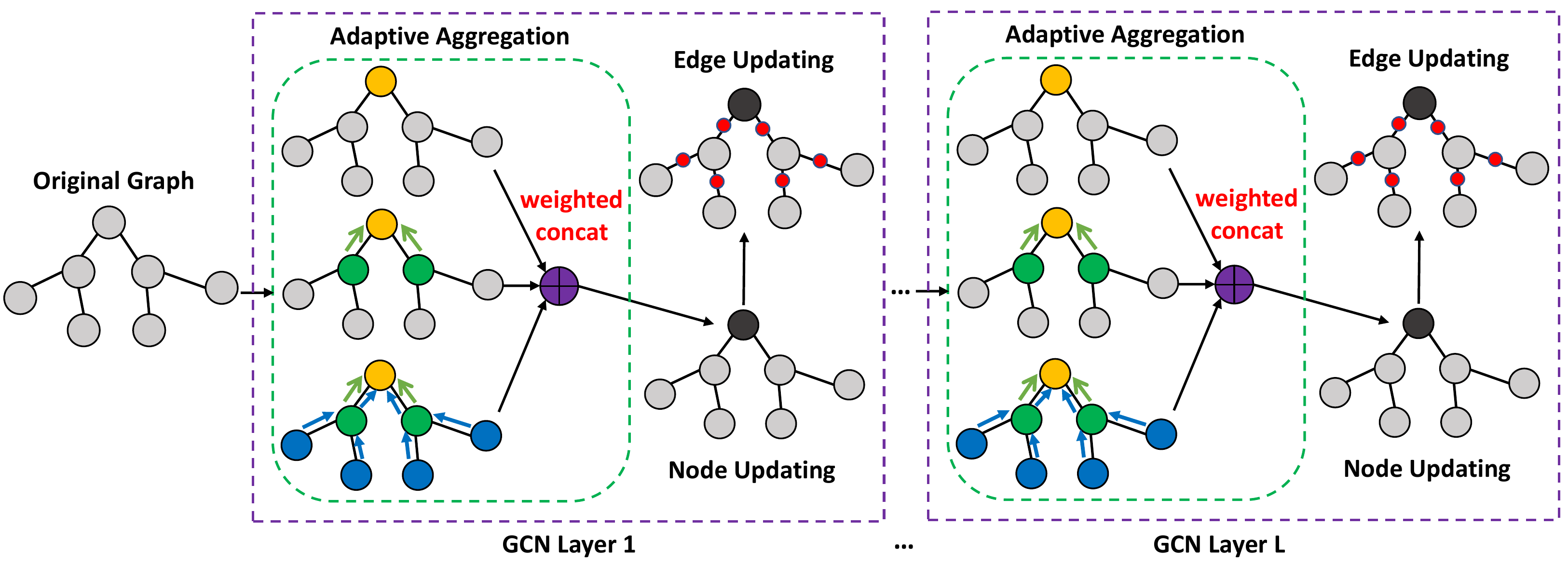}
\vspace{-0.05in}
\caption{Illustration of our adaptive aggregation module. }
\label{fig_aggregation}
\vspace{-0.15in}
\end{figure}

\begin{figure*}[ht]
\vspace{0.03in}
\centering
\includegraphics[width=0.88\textwidth]{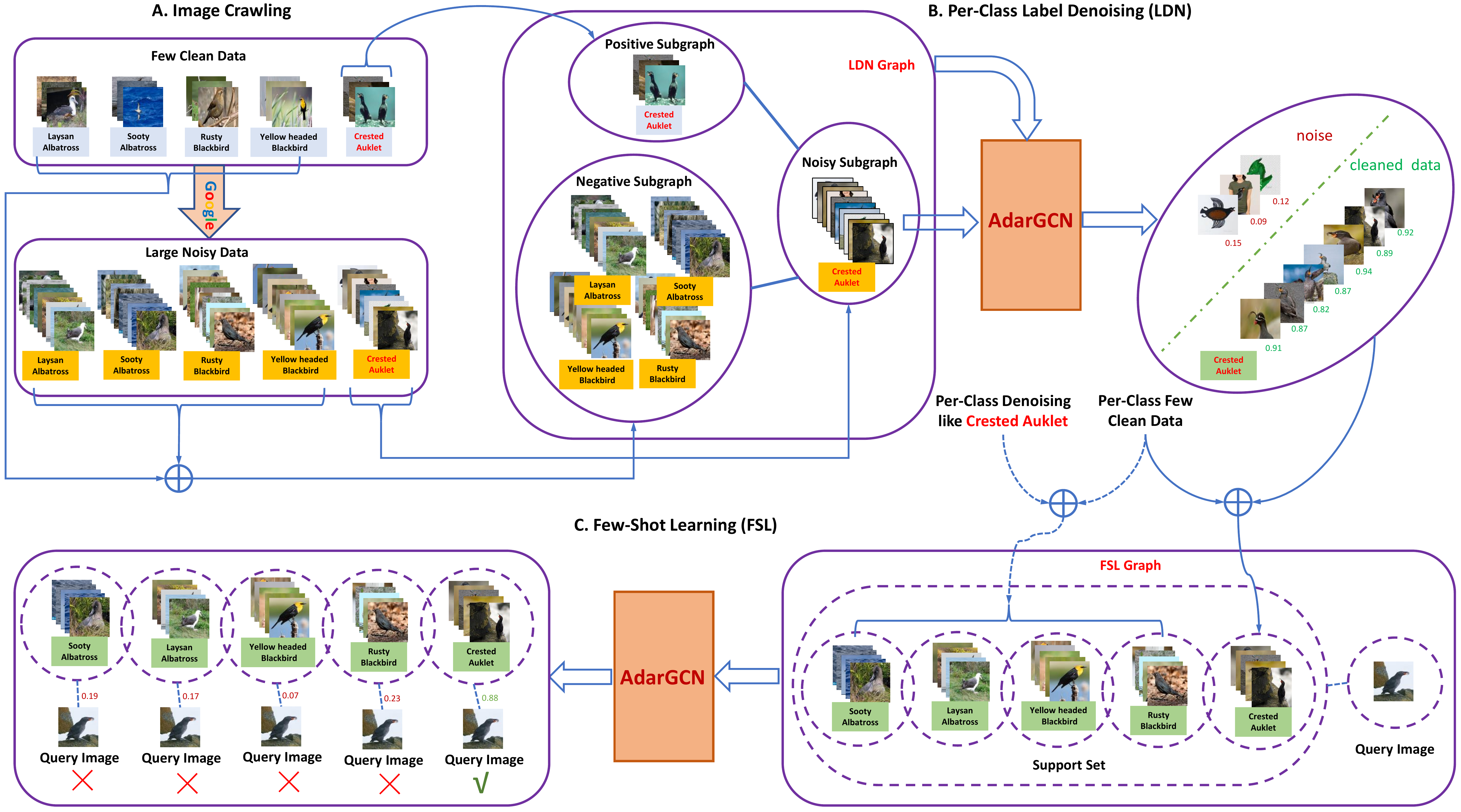}
\caption{Schematic of the proposed AdarGCN model for few-shot few-shot learning (FSFSL). }
\label{fig:total}
\vspace{-0.15in}
\end{figure*}

In this work, we define a new setting termed few-shot few-shot learning (FSFSL), where only few labelled samples from both source and target classes are available for model training. To overcome the source class data scarcity problem under the FSFSL setting, a natural solution would be to crawl sufficient images from the web by searching with the name of each source class (e.g. utilizing Google Image Search). However, although the crawled data contain additional training images, it also inevitably consists of large quantities of irrelevant ones. To fully exploit the crawled noisy images of each source class for FSFSL, label denoising (LDN) is required as a preprocessing step. 

Inspired by the successful use of graph convolutional network (GCN) \cite{kipf2016semi,defferrard2016convolutional,simonovsky2017dynamic,gao2018large} in many vision problems, we focus on GCN-based LDN in this work. Specifically, for each source class, the few clean training images of this source class are used as the positive samples, while the clean and crawled images of the other source classes are used as the negative samples. Given a specific source class, although the crawled images of the other source classes are noisy, it is safe to assume that most if not all of them are negative w.r.t. this source class. This fact is taken advantage of when we design our GCN-based LDN model.  

With the web images cleaned by our GCN-based LDN model, they can be merged with the original clean training images to form an enlarged training set. Our new FSFSL setting thus becomes the conventional one, and any existing FSL methods can be employed here. However, there are still noisy training images undetected by the LDN -- no matter how effective it is, it is not perfect. Consequently, the FSL model must be able to cope with this data noise problem. To this end, we propose a novel GCN-based FSL method to better solve the FSL problem with the augmented noisy training data. Different from previous GCN-based FSL methods \cite{GNN,EGNN,Gidaris_2019_CVPR}, we design an adaptive aggregation GCN (AdarGCN) which can perform adaptive aggregation based on a multi-head multi-level aggregation module (see Fig.~\ref{fig_aggregation}). With our AdarGCN, how much and how far the information carried by each node is propagated to the rest of the graph structure is controlled by each head, making the propagation controllable and adaptive to each instance. An aggregation gate with learnable parameters is then used to dynamically determine the weight of each head when fusing multiple heads. In this way, the negative impact of a noisy training sample can be limited to a small neighborhood of the corresponding node, thus effectively diminishing its detrimental effect. As illustrated in Fig.~\ref{fig:total}, our AdarGCN is used to solve both tasks (i.e.~LDN and FSL) because in both tasks, dealing with noisy training images is the key and our AdarGCN is effective  under both the new and conventional FSL settings. We also empirically observe that with AdarGCN, (1) the GCN can be deeper than existing ones which are typically limited by the depth of layers due to over-smoothing, bringing additional performance gain, and (2) it beats the state-of-the-art alternatives even under the conventional FSL setting, indicating that AdarGCN benefits from dealing with the clean but outlying samples under the conventional FSL.

Our contributions are: (1) We define a new FSL setting termed FSFSL, which is more challenging yet more realistic than the conventional FSL setting. (2) A two-stage solution is provided for FSFSL: 1) crawling sufficient source class training images from the web and performing label denoising on them; 2) solving the FSL problem after merging the cleaned web images with the original training samples. (3) Both the LDN and FSL tasks involved in our FSFSL setting are addressed by proposing a novel GCN model termed AdarGCN. It is different from existing GCN models in that it can perform adaptive aggregation to alleviate the effects of noisy training samples. Extensive experiments show that our AdarGCN achieves state-of-the-art results under both FSL settings. The code and dataset will be released soon. 

\section{Related Work}

\noindent\textbf{Few-Shot Learning}. Meta-learning based methods \cite{Reptile,Mini_split,MAML,ProtoNet,MatchingNet,RelationNet,SNAIL,IMP,lee2019meta,rusu2019meta} have dominated recent FSL research. Apart from metric learning solutions \cite{ProtoNet,MatchingNet,RelationNet,IMP}, another promising approach is learning to optimize \cite{Mini_split,MAML,lee2019meta,rusu2019meta}. More recently, methods based on feature hallucination and synthesis \cite{Hariharan2017iccv,Schwartz2018nips} or predicting parameters of the network \cite{Qiao2018cvpr,Qi2018cvpr} have been developed. However, the promising performance of existing FSL methods is highly dependent on the assumption that there exist sufficient training labelled samples. In this work, we thus focus on a new FSFSL setting (only with a few labelled samples per source class). Even though our AdarGCN is designed for this new setting, it is found to be extremely competitive under the conventional FSL setting (see Table \ref{FSL_comp}). This suggests that the outlying samples problem, largely ignored by existing FSL methods so far, should also be addressed even if the source class data are plenty. 

\noindent\textbf{Graph Convolutional Networks}. GCN is designed to work directly on graphs and leverage their structural information \cite{kipf2016semi, defferrard2016convolutional, simonovsky2017dynamic, gao2018large}. Recently, GCN has been employed in various problems \cite{zhao2019semantic, zhang2019dual, shi2019two, ma2019disentangled, ying2018graph, schlichtkrull2018modeling, levie2018cayleynets, li2018deeper, li2018combinatorial, zhao2019t}. In particular, label denoising with GCN \cite{iscen2019graph, Gidaris_2019_CVPR} has attracted much attention. In \cite{iscen2019graph}, its focus is on fully exploiting sufficient noisily labelled samples from target classes for FSL (which is against the standard FSL setting), and the core transfer problem implied in FSL remains unexplored. To overcome these drawbacks, we choose to study a new FSFSL setting in our current work. In \cite{Gidaris_2019_CVPR}, a GCN-based denoising autoencoder is proposed to generate the classification weights for both source and target classes under generalized FSL \cite{schonfeld2019generalized}, but no GCN-based label denoising problem is concerned in \cite{Gidaris_2019_CVPR}. Moreover, although GCN has been directly used in a number of recent FSL methods \cite{GNN, EGNN, Gidaris_2019_CVPR, iscen2019graph}, our AdarGCN is different in that it can perform adaptive aggregation for FSL and is able to cope with both noisy and outlying training samples. Our results show that the new GCN is clearly better (see Table~\ref{FSL_comp}).

\vspace{-0.1cm}
\section{Methodology}

\vspace{-0.1cm}
\subsection{Problem Definition}

We formally define the few-shot few-shot learning (FSFSL) problem as follows. Let $C_s$ denote a set of source classes and $C_t$ denote a set of target classes ($C_s \bigcap C_t = \emptyset$). We are given a $k_1$-shot sample set $\mathcal{D}_{s}$ from the source classes, a $k$-shot sample set $\mathcal{D}_{t}$ from the target classes, and a test set $\mathcal{T}$ from the target classes. Concretely, the first small sample set can be defined as $\mathcal{D}_{s}=\{(x_i, y_i)|y_i \in \mathcal{C}_s, i=1,...,N_s\}$, where $y_i$ denotes the class label of sample $x_i$ and $N_s$ denotes the number of samples in $\mathcal{D}_{s}$. Since each source class from $\mathcal{D}_{s}$ has only $k_1$ labelled samples, we have $N_s = k_1 |C_s|$. Similarly, the second small sample set can be defined as $\mathcal{D}_{t}=\{(x_i, y_i)|y_i \in \mathcal{C}_t, i=1,...,N_t\}$, where $N_t=k|C_t|$ (each target class has $k$ labelled samples). The goal of FSFSL is thus to train a model with $\mathcal{D}_s$ that can generalize well to $\mathcal{T}$. Note that our new FSFSL problem is clearly more challenging than the conventional FSL problem, since $\mathcal{D}_s$ only has  few samples per class.

As in previous works \cite{MAML,ProtoNet,MatchingNet,RelationNet,SNAIL}, we train a FSL model with $n$-way $k$-shot classification tasks. Concretely, each $n$-way $k$-shot task is defined over a randomly-sampled episode $\{\mathcal{S}_e, \mathcal{Q}_e\}$, where $\mathcal{S}_e$ is the support set having $n$ classes and $k$ samples per class, and $\mathcal{Q}_e$ is the query set. Each episode is sampled as follows: we first sample a small set of source classes $C_e=\{C_i|i=1,...,n\}$ from $C_{s}$, and then generate $\mathcal{S}_e$ and $\mathcal{Q}_e$ by sampling $k$ support samples and $q$ query samples from each class in $C_e$, respectively. Formally, we have $\mathcal{S}_e=\{(x_i, y_i)|y_i\in C_e,i=1,...,n\times k\}$ and $\mathcal{Q}_e=\{(x_i, y_i)|y_i\in C_e, i=1,...,n\times q\}$, where $\mathcal{S}_e \bigcap \mathcal{Q}_e=\emptyset$. A FSL model is then trained by minimizing the gap between its predicted labels and the ground-truth labels over the query set $\mathcal{Q}_e$.

\subsection{Two-Stage Solution}

To overcome the lack of training samples from source classes in FSFSL, we provide a two-stage solution, as shown in Figure~\ref{fig:total}. The first stage consists of image crawling and GCN-based LDN, as shown in Figure~\ref{fig:total}(A) and Figure~\ref{fig:total}(B). For each source class $c\in C_s$, we only have a small set of $k_1$ clean images initially: $X^s_c = \{x_i|(x_i,y_i)\in \mathcal{D}_s, y_i=c, i=1,...,N_s\}$. To augment the small set $X^s_c$, we then crawl another set of $k_2$ additional images from the web by image searching with the name of source class $c\in C_s$: $X^{web}_c=\{x_i|i=1,...,k_2\}$. As expected,  there exists much noise in $X^{web}_c$. Therefore, we propose a GCN-based LDN method to reduce the noise in $X^{web}_c$ and obtain a set of cleaned images $X^{d}_c \subset X^{web}_c$. We then define the set of denoised samples as: $\mathcal{D}_d = \{(x,y)|x\in X^{d}_c, y=c, c=1,...,|C_s|\}$. Moreover, the second stage consists of GCN-based FSL, as shown in Figure~\ref{fig:total}(C). We leverage both $\mathcal{D}_s$ and $\mathcal{D}_d$ to train our GCN-based FSL model. For both the LDN and FSL tasks, we design an adaptive aggregation GCN (AdarGCN) model (see Figure~\ref{fig:GNN_struct}).

\subsection{GCN-Based LDN}

In the first stage, we perform GCN-based LDN over the noisy images crawled for each source class. Specifically, given a source class $c \in C_{s}$, we have a positive image set $X^{+}_c=X^{s}_c$, a noisy image set $X^{*}_c=X^{web}_c$, and a negative image set $X^{-}_c=\{X^{+}_i\cup X^{*}_i|i \in C_s, i \ne c\}$, as shown in Fig.~\ref{fig:total}(B). Before per-class LDN, we pretrain a simple embedding network (e.g. four-block ConvNet) on $X^{+}_c$ like ProtoNet \cite{ProtoNet} to extract $d$-dimensional image feature vectors, which is consistent with the second stage (the same simple backbone is used for GCN-based FSL).

To construct an LDN graph for each source class $c \in C_{s}$, we generate a mini-batch by randomly selecting $m^+$ images from $X^{+}_c$, $m^*$ images from $X^{*}_c$, and $m^-$ images from $X^{-}_c$ (see Fig.~\ref{fig:total}(B)). The image feature matrices of these three groups of samples are respectively denoted as $V_c^{+} \in \mathbb{R}^{m^{+} \times d}$, $V_c^{*} \in \mathbb{R}^{m^{*} \times d}$, and $V_c^{-} \in \mathbb{R}^{m^{-} \times d}$. The node feature matrix is thus defined as $V_c=[V_c^{+}; V_c^{*}; V_c^{-}]=[v_1; \cdots; v_M] \in \mathbb{R}^{M \times d}$, where $M = m^{+}+m^{*}+m^{-}$. The initial symmetric adjacency matrix $A_c=\{a_{ij}\} \in \mathbb{R}^{M \times M}$ is defined as: $a_{ij}=0$ if $v_i$ and $v_j$ respectively come from $V_c^{+}$ and $V_c^{-}$, and $a_{ij}=1$ otherwise (see Fig.~\ref{fig:total}(B)). This choice of constructing $A_c$ ensures that the positive and negative samples cannot be directly confused by each other.

We denote the above LDN graph as $\mathcal{G}_c=(V_c,E_c)$, where the edge feature matrix $E_c = \{e_{ij}\} \in \mathbb{R}^{M \times M}$. In this work, we exploit a 3-layer AdarGCN model for label denoising, which is defined in Section~\ref{sec:Adar}. Specifically, to adapt AdarGCN to the LDN task, we make four modifications to its architecture: (1) For each edge updating (EU) unit, we set $e_{ij}=0$ when $a_{ij}=0$, after the sigmoid operation to avoid direct label confusion during label propagation. (2) Before the node updating (NU) unit of the first GCN layer, we perform EU to initialize $E_c$. (3) For each NU unit, we perform a linear transformation after the concat operation. (4) For the last GCN layer, we drop the EU unit and use a sigmoid function to output the predicted score for each sample.

Note that our GCN-based LDN model can be regarded as a binary classifier, outputting 1 for positive samples and 0 for negative ones. However, unlike the traditional image classification, our model can make full use of uncertain samples (i.e. images from $X^{*}_c$) by aggregating similar nodes for more effective label propagation. Let $\hat{y}_i$ be the predicted score of each sample $x_i$ ($i=1,...,M$). The loss for GCN-based LDN is defined as follows:
\begin{equation}
\mathcal{L}_{LDN} = -\frac{1}{m^+}\sum_{i=1}^{m^+}\log(\hat{y}_i)-\frac{1}{m^-}\hspace{-0.1in} \sum_{i=M-m^- + 1}^M \hspace{-0.12in}\log(1-\hat{y}_i).
\label{pass_label_loss}
\end{equation}
Although our GCN-based LDN model ignores the direct back-propagation w.r.t. the loss of noisy images (whose labels are uncertain), it can learn better representation for uncertain images by aggregating both certain and uncertain images, followed by back-propagation w.r.t. the loss of certain images. Notably, our GCN-based LDN model is shown to outperform multi-layer perceptron (MLP) which copes with each sample independently (see Table~\ref{LDN_comp_both}).

After the GCN training process, since each uncertain sample $x \in X^{*}_c$ appears in multiple LDN graphs w.r.t. source class $c$, we average the obtained multiple predicted scores as the probability of being positive for each sample $x$. If this probability is greater than a preset threshold, we then add $(x,c)$ to the set of denoised samples $\mathcal{D}_d$.

\subsection{GCN-Based FSL}

In the second stage, given $\mathcal{D}_s\cup\mathcal{D}_d$, we train our GCN-based FSL model by episodic sampling. For each episode, we randomly select $n\times k$ samples to form the support set $\mathcal{S}_e$ and $n\times q$ samples to form the query set $\mathcal{Q}_e$. The embedding network $f_\varphi$ is trained jointly with the GCN module to obtain the feature representations of all samples from $\mathcal{S}_e \cup \mathcal{Q}_e$: $v_i=f_\varphi(x_i)$, $i=1,...,n\times(k+q)$. Although both transductive (with all query images in one trial) and non-transductive (with a single query image per trial) test strategies are followed in \cite{EGNN}, we only adopt the non-transductive strategy for fair comparison, since most of the state-of-the-art FSL methods are non-transductive. Concretely, for each episode, we construct $n\times q$ graphs, each of which is defined over $n\times k$ support samples and 1 query sample.

We collect the above $n\times q$ graphs as $\mathcal{G} = \{\mathcal{G}_t=(V_t, E _t)|t=1,...,n\times q\}$. For a single graph $\mathcal{G}_t$ in $\mathcal{G}$, $V_t=\{\overbrace{v_1^s, \cdots, v_{n\times k}^s}^{\mathcal{S}_e}, \overbrace{v_t^q}^{\mathcal{Q}_e}\}$. Concisely, we take it as $V_t=\{v_i\}_{i=1}^{n\times k+1}$, along with $E_t=\{e_{ij}\}_{i,j=1,\cdots,n\times k+1}$, where $v_i$ is the node feature obtained by the embedding network, $v_{n\times k+1}$ denotes the node feature of the query image, and $e_{ij}$ is the edge feature w.r.t. $v_i$ and $v_j$. For $v_i, v_j\in \mathcal{S}_e$, $e_{ij}=1$ if $v_i$ and $v_j$ come from the same class and $e_{ij}=0$ otherwise. When $v_i\notin \mathcal{S}_e$ or $v_j\notin \mathcal{S}_e$, we also set $e_{ij}=0$ due to the unknown label of $v_i$ or $v_j$.

\begin{figure}[t]
\centering
\includegraphics[width=0.93\columnwidth]{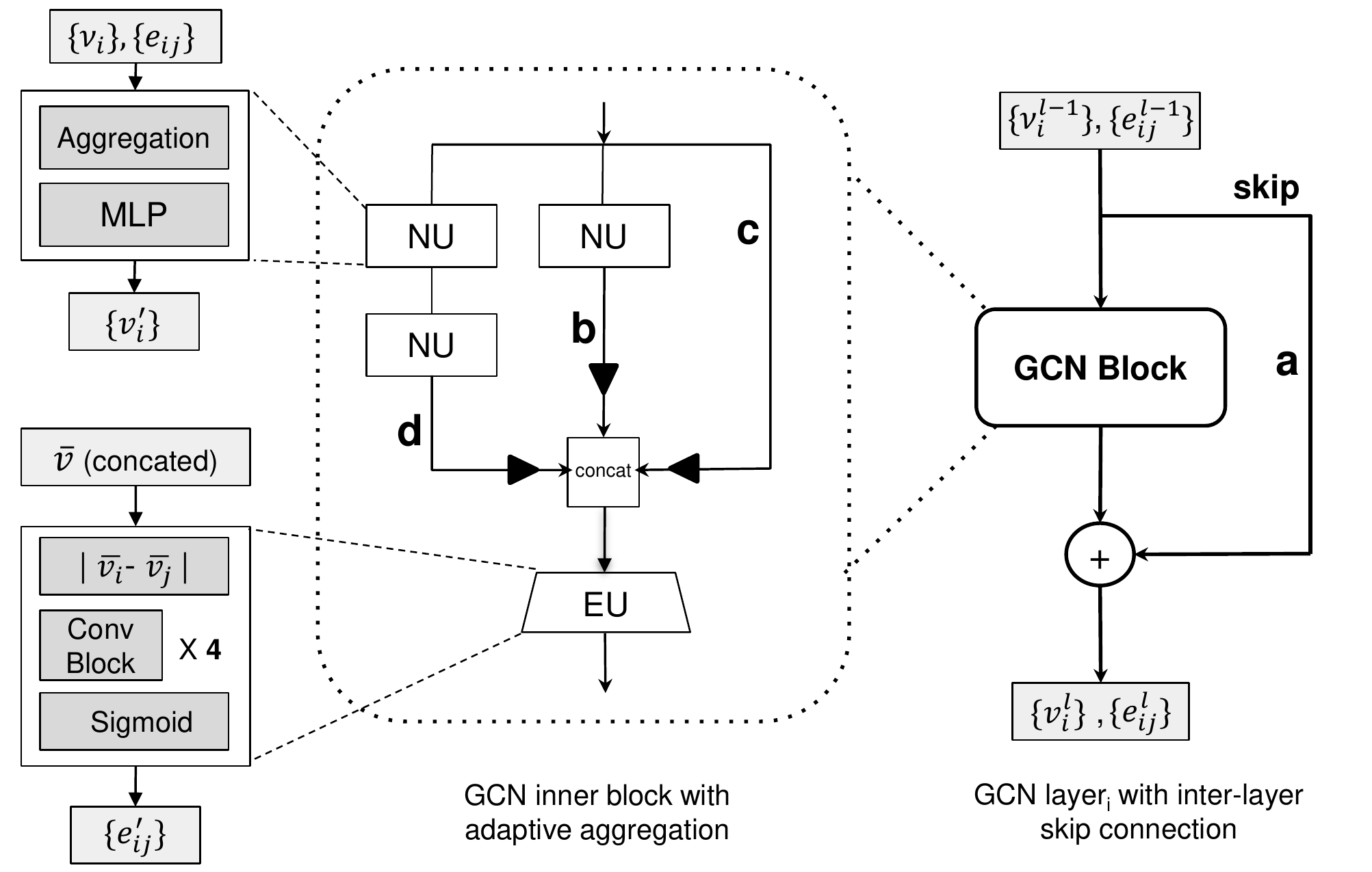}
\caption{Illustration of the network architecture of our AdarGCN model. Notations: NU -- node updating; EU -- edge updating. }
\label{fig:GNN_struct}
\vspace{-0.12in}
\end{figure}

The full GCN-based FSL model is stacked by $L$ GCN layers with the same AdarGCN   architecture shown in Figure~\ref{fig:GNN_struct}. In this work, we set $L=3$. Given a graph $\mathcal{G}_t$ ($t=1,...,n\times q$), the inputs of the first GCN layer (i.e. node feature matrix $V_t^0$ and edge feature matrix $E_t^0$) are obtained in the way mentioned above (where we set $l=0$). For the $l$-th GCN layer ($l=1,...,L$), the inputs $V_t^{l-1}$ and $E_t^{l-1}$ (from the previous layer) are updated to $V_t^{l}$ and $E_t^{l}$.

For GCN training, we choose the binary cross-entropy loss between the ground-truth edge matrix $E^{gt}=\{e_{ij}^{gt}\}_{i,j=1,\cdots,n\times k+1}$ and the edge feature matrices of all $L$ GCN layers $\{E_t^l=\{e^l_{ij}\}_{i,j=1,\cdots,n\times k+1}\}_{l=1}^L$, where $e_{ij}^{gt}=1$ if $x_i$ and $x_j$ come from the same class and $e_{ij}^{gt}=0$ otherwise. Formally, for each graph $\mathcal{G}_t$, the binary cross-entropy loss of the $l$-th GCN layer is defined as:
\begin{equation}
\mathcal{L}_l^{t}= \hspace{-0.03in}-\hspace{-0.07in}\sum_{i=1}^{n\times k+1}\sum_{j\not=i} e_{ij}^{gt}\cdot\log(e_{ij}^l) + (1-e_{ij}^{gt})\cdot\log(1-e_{ij}^l).
\label{bi_loss}
\end{equation}
Taking all graphs (each for a query image) and all GCN layers on board, we compute the overall cross-entropy loss for a training episode as follows:
\begin{equation}
    \begin{aligned}
    \mathcal{L}_{FSL} = \sum_{t=1}^{n\times q}\sum_{l=1}^{L}\lambda ^l\cdot \mathcal{L}_l^{t},
    \end{aligned}
    \label{total_loss}
\end{equation}
where $\lambda ^l$ denotes the loss weight of the $l$-th GCN layer.

For GCN inference, the edge feature matrix $E_t^L$ of the last GCN layer can be used to predict the label of the unique query image. Concretely, the predicted scores of the query image are collected as $\hat{y} = \{\hat{y}_1, \cdots, \hat{y}_{n\times k}\}$, where $\hat{y}_i=e^L_{n\times k+1, i}~(0\leq\hat{y}_i\leq1)$, being the predicted probability of the query image coming from the class that support sample $x_i$ belongs to ($i=1,...,n\times k$). The classification probability of the query image is:
\begin{equation}
p_c  = \sum_{i=1}^{n\times k} \mathbbm{1}(y_i=c)\cdot \hat{y}_i/\sum_{i=1}^{n\times k} \hat{y}_i,
\label{infer_prob}
\end{equation}
where $\mathbbm{1}$ denotes the indicator function, $y_i$ denotes the class label of support sample $x_i$, and $c$ denotes the $c$-th class label in the episode ($c=1,...,n$).

\subsection{Network Architecture of AdarGCN}
\label{sec:Adar}

For both LDN and FSL tasks involved in our FSFSL setting, we design an AdarGCN model, as illustrated in Figure~\ref{fig:GNN_struct}. Different from existing GCN models \cite{GNN,EGNN,Gidaris_2019_CVPR,kipf2016semi,li2019can}), our AdarGCN induces adaptive aggregation into GCN training to better control the information propagation from each node to the rest to the graph structure. 

Formally, for the $l$-th GCN layer ($l=1,...,L$) of our AdarGCN, given the node feature matrix $V^{l-1}$ and the edge feature matrix $E^{l-1}$ as inputs, the output $V^l$ can be obtained by adding the node feature matrix from the GCN inner block and that from the previous GCN layer:
\begin{equation}
 V^l = V^{l-1} + V_{block},
\end{equation}
which is essentially implemented by the inter-layer skip-connection branch \textbf{a} (see Figure~\ref{fig:GNN_struct}).

Within the GCN inner block of the $l$-th GCN layer, we design a multi-head multi-level aggregation module to aggregate the node features adaptively with different aggregation complexities, as illustrated in Figure~\ref{fig:GNN_struct}. For this GCN inner block, we choose to update the node feature matrix and edge feature matrix successively.

Specifically, node feature updating is achieved by the adaptive aggregation among the three updating branches \textbf{c, b, d} with different degrees of aggregation. Branches \textbf{c, b, d} update the node features respectively with $0,1,2$ iterations: one iteration update is denoted by a Node Updating (NU) unit which consists of an aggregation module and a MLP module. The outputs of branches \textbf{c, b, d} are given by:
\begin{equation}
\begin{aligned}
V_c^l & = V^{l-1}, \\
V_b^l & = f_{\theta_{b}}(E^{l-1}\cdot V^{l-1}),\\
V_d^l & = f_{\theta_{d1}}(E^{l-1}\cdot f_{\theta_{d2}}(E^{l-1}\cdot V^{l-1})),
\end{aligned}
\end{equation}
where $\theta_{b}$ collects the parameters of the MLP module in branch \textbf{b}, while $\theta_{d1}$ and $\theta_{d2}$ collect the parameters of the two MLP modules in branch \textbf{d}, respectively. As illustrated in Figure \ref{fig_aggregation}, how far information can travel along each head/branch differs -- \textbf{d} has the farthest influence whilst \textbf{c} the shortest (each node itself). For adaptive aggregation, the adaptive weight of each of branches \textbf{c, b, d} is computed with a fully connected (FC) layer:
\begin{equation}
 w_c = \textrm{FC}(V_c^l),~w_b = \textrm{FC}(V_b^l),~w_d = \textrm{FC}(V_d^l),
\end{equation}
where $\textrm{FC}(\cdot)$ denotes the output of a FC layer, followed by a sigmoid function. The total node update within the GCN inner block is formulated as:
\begin{equation}
 \Bar{V}^l = \mathrm{concat}(w_c\cdot V_c^l,~w_b\cdot V_b^l,~w_d\cdot V_d^l).
\end{equation}
Note that more than three branches can be employed here, but empirically we found that more branches leads to no further gains. We thus use only three branches in this work.

For edge feature updating, we denote it with an Edge Updating (EU) unit. EU aims to learn the distance metric given node features as inputs, which includes a distance computing operation, 4 conv blocks, and a sigmoid activation function, as shown in Figure~\ref{fig:GNN_struct}.

\section{Experiments}

\subsection{New FSFSL}

\subsubsection{Datasets and Settings}
\label{sect:fsfsl_setting}

\noindent\textbf{Datasets}. Two benchmark datasets are selected: (1) \textbf{mini-ImageNet}: This dataset is proposed in \cite{MatchingNet} and derived from ILSVRC-12 \cite{Imagenet}. It consists of 100 classes totally. As in \cite{Mini_split}, this dataset is split into 64 training classes, 16 validation classes, and 20 test classes. Each image is resized to 84${\times}$84. (2) \textbf{CUB}: The CUB dataset \cite{CUB-200-2011} is particularly suitable for our new FSFSL setting. Concretely, since the number of images per class is less than 60, the FSL problem on CUB is essentially a FSFSL problem. Although CUB has widely used for FSL, this work is the first to identify the problem and to provide a solution. This dataset consists of 200 bird species totally. We split CUB into 100 training classes, 50 validation classes, and 50 test classes. Each image is also resized to 84${\times}$84.

\noindent\textbf{FSFSL Settings}. Let $k_1$ be the number of original clean images per training class (i.e. source class), and $k_2$ be the number of crawled noisy images per training class. In this work, we set $k_1$=10, 20, or 50, and $k_2$=1,200. As in the state-of-the-art works on GCN-based FSL \cite{GNN,EGNN}, the four-block ConvNet network is used as the embedding network. For both LDN and FSL tasks involved in our new FSFSL setting, the same embedding network is used. As in \cite{GNN,EGNN}, the 5-way 5-shot accuracy is computed over 600 episodes randomly sampled from the test set: each test episode have 5 support images and 15 query images per class. Although both transductive and non-transductive test strategies are followed in \cite{EGNN}, we only take the non-transductive test strategy on board for fair comparison, since most of the state-of-the-art FSL methods are non-transductive.

\noindent\textbf{Implementation Details}. (1) \textbf{GCN-Based LDN}: The four-block ConvNet network pretrained on the training set is used as the feature extractor. The dimensionality of the output features is 128. For GCN training over each training class, a mini-batch consists of three types of images from this class: 5 positive images, 5 negative images, and 50 crawled noisy images\footnote{Out of these crawled images, around 40\% are noise. After LDN using our AdaGCN, this percent is reduced to around 10\%. Some examples of the removed images can be found in the suppl. material.} (see  Figure~\ref{fig:total}). We construct an LDN graph over each mini-batch. We set a learning rate of ${0.001}$, a dropout probability of ${0.5}$, and a mini-batch size of 8. We adopt the Adam optimizer \cite{Adam}, with totally 500 training epochs. For each training class, we select the denoised images with prediction scores $\ge 0.5$ for the subsequent GCN-based FSL task. (2) \textbf{GCN-Based FSL}: With the same 4-block embedding network, our GCN model for FSL is trained by the Adam optimizer \cite{Adam} with a initial learning rate of ${0.001}$ and a weight decay of ${1e-6}$. We also use label smoothing as in \cite{lee2019meta}. During the training phase, we cut the learning rate in half every 10,000 episodes and set total training episodes as 50,000. In the 5-way 5-shot scenario, each mini-batch has 32 training episodes, and each episode consists of 25 support images and 5 query images (5 shot support samples and 1 query sample per class). Within a training episode, we construct a graph over 25 support images and 1 query image for each query image.

\noindent\textbf{Compared Methods}. Since our FSFSL setting includes both LDN and FSL tasks, we compare our AdarGCN-LDN and AdarGCN-FSL with various LDN and FSL alternatives, respectively. When comparing our AdarGCN-LDN with other LDN methods including label propagation (LP) \cite{zhou2004learning,zhu2003semi}, MLP, GCN \cite{kipf2016semi} and ResGCN \cite{li2019can}, we employ the same subsequent FSL model (i.e. AdarGCN-FSL) for fair comparison. Note that a score threshold of 0.5 is selected for all LDN methods except LP to classify the positive and negative samples. For LP-based LDN, a score threshold of 0 is used otherwise, since the positive samples are labelled as `1' and the negative samples as `-1'. Similarly, when comparing our AdarGCN-FSL with various FSL baselines, we adopt the same LDN model (i.e. LDN-AdarGCN) for fair comparison. We focus on representative/state-of-the-art FSL methods including MatchingNet \cite{MatchingNet}, MAML \cite{MAML}, ProtoNet \cite{ProtoNet}, IMP \cite{IMP}, Baseline++ \cite{Chen2019ICLR}, GCN \cite{GNN}, wDAE-GNN \cite{Gidaris_2019_CVPR}, and EGCN \cite{EGNN}.

\begin{table}[t] 
\vspace{0.05in}
\centering
\tabcolsep5pt
\scalebox{0.92}{
\begin{tabular}{l|c|c|c}
\hline
Method & $k_1$=10 & $k_1$=20 & $k_1$=50 \\
\hline
FSL (w/o crawled noisy images) & 40.91  & 50.01 & 55.04\\
FSL (w/ crawled noisy images) & 59.20  & 59.37 & 59.64\\
\hline
FSL+LDN (LP) &  60.21  &  62.83  &  64.74 \\
FSL+LDN (MLP) &  60.19  &  62.88  &  64.25\\
FSL+LDN (GCN \cite{kipf2016semi})& 61.22  &  63.27  &  65.36\\
FSL+LDN (ResGCN \cite{li2019can})  & 61.48  & 63.79  &  65.92 \\
FSL+LDN (ours) &  \bf63.37  &  \bf65.12  &  \bf66.85 \\
\hline
\end{tabular}
}
\vspace{0.03in}
\caption{Comparative results by various label denoising (LDN) methods under the new FSFSL setting on mini-ImageNet. FSL denotes our GCN-based FSL method with our AdarGCN model. }
\label{LDN_comp_mini}
\vspace{-0.0in}
\end{table}

\begin{table}[t]  
\vspace{0.05in}
\centering
\tabcolsep5pt
\scalebox{0.92}{
\begin{tabular}{l|c|c|c}
\hline
Method & $k_1$=10 & $k_1$=20 & $k_1$=50 \\
\hline
FSL (w/o crawled noisy images) & 58.89  & 68.33 & 76.16\\
FSL (w/ crawled noisy images) & 76.35  &  76.83  &  77.18 \\
\hline
FSL+LDN (LP)   &  76.94  &  77.98  &  78.87 \\
FSL+LDN (MLP)  &  77.10  &  78.06  &  78.92 \\
FSL+LDN (GCN \cite{kipf2016semi})   &  77.44  &  78.56   &  79.32 \\
FSL+LDN (ResGCN \cite{li2019can}) &  77.52   &  78.69  &   79.69\\
FSL+LDN (ours)   &  \bf79.16  &  \bf79.82  &   \bf80.88\\
\hline
\end{tabular}
}
\vspace{0.03in}
\caption{Comparative results by various label denoising (LDN) methods under the new FSFSL setting on CUB. FSL denotes GCN-based FSL with our AdarGCN model. }
\label{LDN_comp_cub}
\vspace{-0.0in}
\end{table}

\vspace{-0.1cm}
\subsubsection{Comparison to LDN Alternatives}

The comparative results for the label denoising task on the two datasets are shown in Tables~\ref{LDN_comp_mini} and \ref{LDN_comp_cub}, respectively. It can be seen that: (1) Adding the crawled images (although noisy) to the original few clean training data leads to consistent and significant improvements for different values of $k_1$. The improvements are particularly salient when $k_1$ takes a smaller value. (2) All LDN methods can further improve the FSL performance (see `FSL+LDN' vs. `FSL (w/ crawled noisy images)') by imposing label denoising over the crawled images, showing the effectiveness of LDN under our new FSFSL setting. (3) Our AdarGCN-LDN achieves the best label denoising results among all LDN methods. This suggests that GCN is suitable for label denoising, and adaptive aggregation included in our GCN model (i.e. AdarGCN) yields further improvements.

\begin{table}[t] 
\vspace{0.05in}
\centering
\tabcolsep5pt
\scalebox{0.95}{
\begin{tabular}{l|c|c}
\hline
Method & mini-ImageNet & CUB \\
\hline
LDN+FSL (MatchingNet \cite{MatchingNet}) &51.59 & 66.12\\
LDN+FSL (MAML \cite{MAML}) & 59.67 & 73.50\\
LDN+FSL (ProtoNet \cite{ProtoNet}) &  64.73  &  74.48 \\
LDN+FSL (IMP \cite{IMP}) & 65.44 & 78.61\\
LDN+FSL (Baseline++ \cite{Chen2019ICLR}) & 64.55 & 77.90\\
LDN+FSL (GCN \cite{GNN}) & 64.80 & 74.59\\
LDN+FSL (wDAE-GNN \cite{Gidaris_2019_CVPR}) &  63.26  &  74.23 \\
LDN+FSL (EGCN \cite{EGNN}) &  65.12  & 78.10\\
\hline
LDN+FSL (ours) &  \bf66.85   &  \bf80.88\\
\hline
\end{tabular}
}
\vspace{0.03in}
\caption{Comparative results by various FSL methods under the new FSFSL setting ($k_1$=50, $k_2$=1,200) on the two datasets. LDN denotes GCN-based LDN with our AdarGCN model.}
\label{FSFSL_comp}
\vspace{-0.05in}
\end{table}

\vspace{-0.0cm}
\subsubsection{Comparison to FSL Alternatives}

The comparative results for the FSL task on the two datasets are shown in Table~\ref{FSFSL_comp}. For fair comparison, all compared methods make use of the same set of denoised samples obtained by our AdarGCN-LDN method. We have the following observations: (1) Our AdarGCN-FSL method performs the best among all FSL methods, validating the effectiveness of our AdarGCN for solving the FSL task. (2) Our method clearly outperforms the latest GCN-based FSL methods \cite{GNN,EGNN,Gidaris_2019_CVPR}, which suggests that adaptive aggregation indeed plays an important role when applying GCN to FSL. (3) Our method also clearly leads to improvements over the state-of-the-art FSL baselines \cite{IMP,Chen2019ICLR}, showing that AdarGCN is a promising model to solve the FSL task.

\begin{table}[t]
\vspace{0.05in}
\centering
\tabcolsep10.5pt
\begin{tabular}{l|c|c}
\hline
GCN Model & LDN & FSL \\
\hline
AdarGCN (branch: b) &  64.36   &  63.13\\
AdarGCN (branches: a, b) &  65.92 & 65.01\\
AdarGCN (branches: a, b, c) &  66.10   &  65.88\\
AdarGCN (branches: a, b, c, d) & \bf66.85  & \bf66.85\\
\hline
\end{tabular}
\vspace{0.03in}
\caption{Ablative results for our AdarGCN model on both LDN and FSL tasks involved in our new FSFSL setting ($k_1$=50, $k_2$=1,200) over mini-ImageNet. }
\label{FSFSL_abl}
\vspace{-0.05in}
\end{table}

\begin{figure}[t]
\centering
\includegraphics[width=0.88\linewidth]{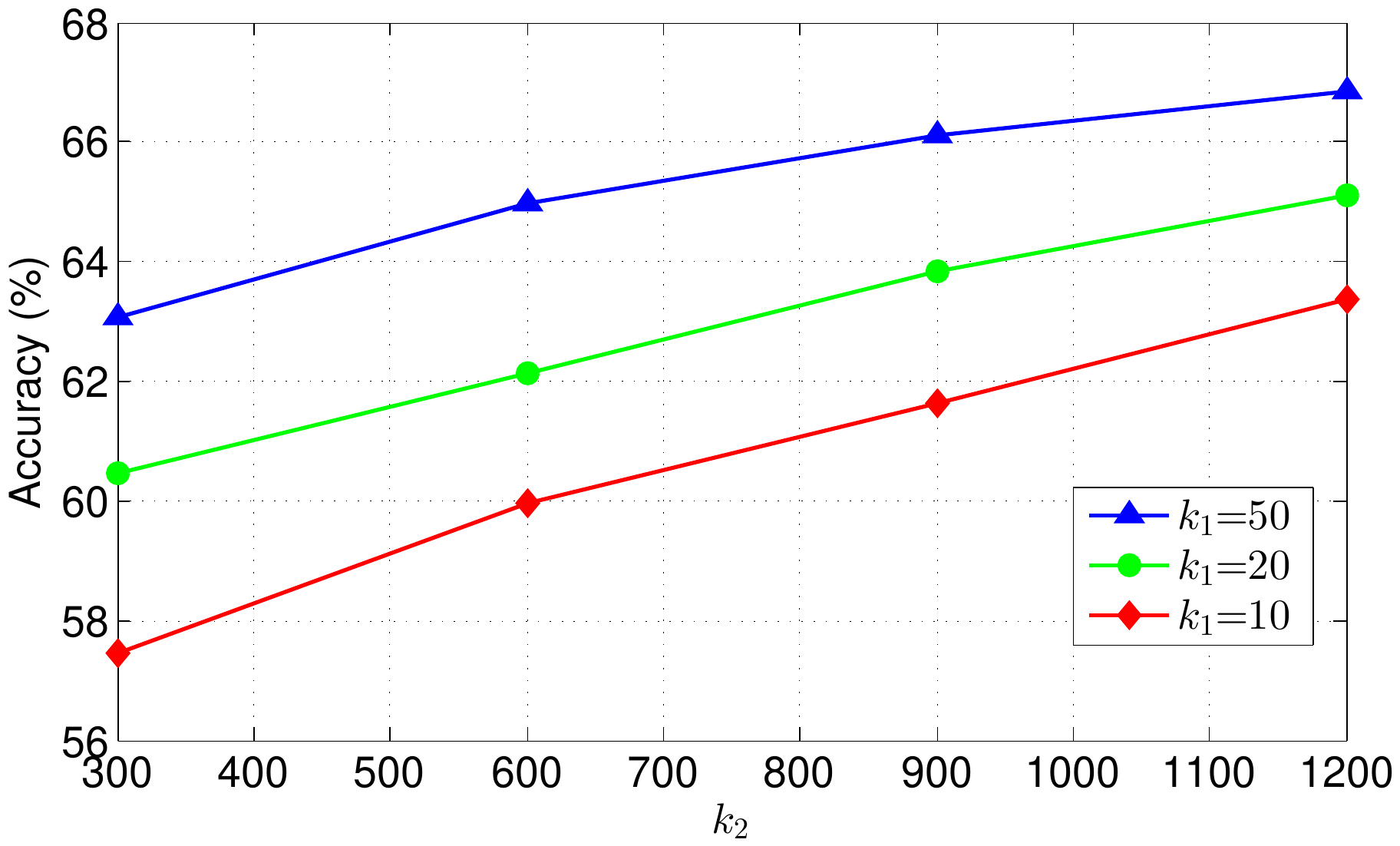}
\caption{Illustration of the effect of different values of $k_2$ on our AdarGCN-LDN method over mini-ImageNet. }
\label{fig_further1}
\vspace{-0.05in}
\end{figure}

\subsubsection{Further Evaluations}

\noindent\textbf{Ablation Study Results}. The ablative results for our AdarGCN model on both the LDN and FSL tasks involved in our new FSFSL setting ($k_1$=50, $k_2$=1,200) are presented in Table~\ref{FSFSL_abl}. Note that AdarGCN-FSL is used for the ablation study on the LDN task, while AdarGCN-LDN is used for the ablation study on the FSL task. We can observe from Table~\ref{FSFSL_abl} that adding more branches leads to more performance improvements on both the LDN and FSL tasks, consistently demonstrating the contribution of each branch (a, b, c, or d in Figure~\ref{fig:GNN_struct}) in our AdarGCN model.

\noindent\textbf{Effect of Different Values of $k_2$}. To show the effect of different values of $k_2$ on our AdarGCN-LDN method, we choose to gradually reduce $k_2$ from 1,200 to 300, and then evaluate the obtained LDN results by forwarding them to the subsequent FSL task (where our AdarGCN-FSL is used). The results in Figure~\ref{fig_further1} show that our AdarGCN-LDN method suffers from gradual performance degradation when $k_2$ decreases from 1,200 to 300. This is essentially consistent with the characteristic of image search engine (i.e. Google): when less relevant images are returned for each source class, there exist less images that truly belong to this source class, resulting in that less denoised training samples can be obtained for the subsequent FSL task (and thus performance degradation is caused).

\begin{figure}[t]
\centering
\includegraphics[width=0.85\linewidth]{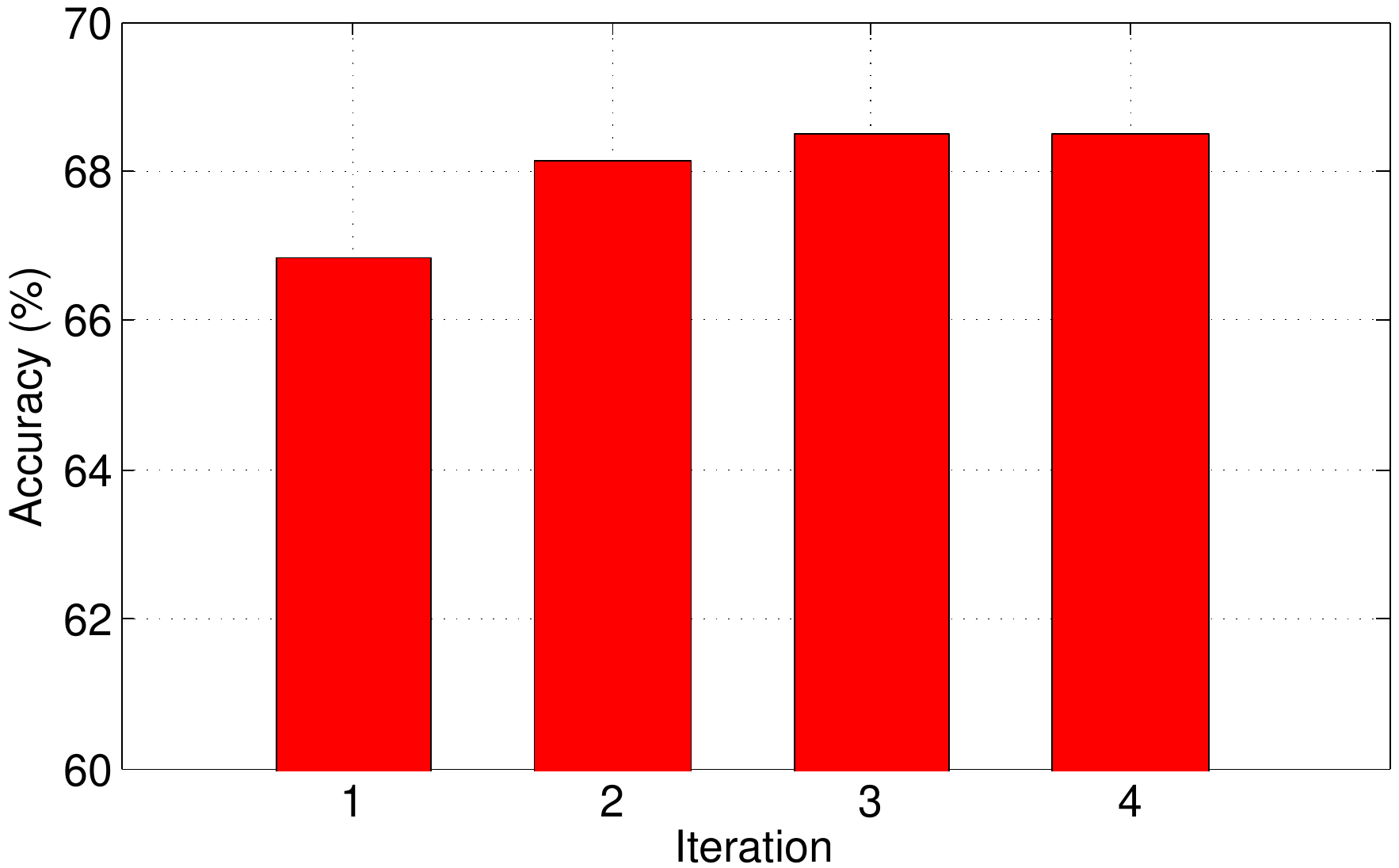}
\caption{Illustration of the effect of iterative optimization on our AdarGCN-LDN method ($k_1$=50, $k_2$=1,200) over mini-ImageNet. }
\label{fig_further2}
\vspace{-0.05in}
\end{figure}

\noindent\textbf{Iterative Optimization for GCN-Based LDN}. Note that the denoised samples obtained by our AdarGCN-LDN method can be easily exploited for another round of GCN-based LDN. In this work, for computational efficiency, we have ignored such iterative optimization in all of the above experiments. To show the effect of iterative optimization on our AdarGCN-LDN method, we present the results obtained by iterative optimization in Figure~\ref{fig_further2}. We can observe that our AdarGCN-LDN method consistently achieves more improvements when more rounds of GCN-based LDN are included and becomes stable after three iterations. 
 
\subsection{Conventional FSL}

\subsubsection{Datasets and Settings}

We further evaluate our AdarGCN-FSL method under the conventional FSL setting. The full mini-ImageNet and CUB datasets are selected for performance evaluation, where mini-ImageNet has 600 samples per class and CUB has less than 60 samples per class. The non-transductive 5-way 5-shot test strategy is adopted, exactly the same as the test strategy used for our new FSFSL setting. Moreover, the implementation details for GCN training remain largely unchanged compared to those described in Section~\ref{sect:fsfsl_setting}. One exception is that: since the number of training samples in the CUB dataset is relatively small, we cut the learning rate in half every 5,000 episodes and set the total number of training episodes as 20,000 on CUB for better optimization.

\begin{table}[t]
\vspace{0.05in}
\centering
\tabcolsep9pt
\scalebox{0.95}{
\begin{tabular}{l|c|c}
\hline
Models & mini-ImageNet & CUB \\
\hline
MatchingNet \cite{MatchingNet} & 55.30 & 68.71\\
ProtoNet \cite{ProtoNet} & 65.77 & 74.70\\
Meta-Learn LSTM \cite{Mini_split} & 60.20 & -- \\
Reptile \cite{Reptile} & 62.74 &  -- \\
MAML \cite{MAML} & 63.11 & 71.33\\
Relation Net \cite{RelationNet} & 67.07 & 69.66 \\
PPA \cite{Qiao2018cvpr} & 67.87  & --  \\
TPN \cite{Liu_2019_ICLR} & 69.86 & -- \\
Shot-Free Meta \cite{Ravichandran_2019_ICCV} & 65.73 & -- \\
R2-D2 \cite{bertinetto2019meta} & 68.40 & -- \\
IMP \cite{IMP} & 68.10 & 71.87  \\
Baseline++ \cite{Chen2019ICLR} & 66.43 & 75.39$^{\dagger}$\\
MetaOptNet \cite{lee2019meta} & 69.51 & 77.10 \\
GCN \cite{GNN} &  66.41 & 74.07\\
wDAE-GNN \cite{Gidaris_2019_CVPR}  &  65.91  &  73.85  \\
EGCN \cite{EGNN} & 66.85 & 74.58 \\
\hline
AdarGCN (ours) & \bf71.48$^{\ddagger}$  & \bf78.04\\
\hline
\end{tabular}
}
\vspace{0.03in}
\caption{Comparative results under the conventional FSL setting. ${\dagger}$ denotes that the result is reproduced since our data split of CUB is different from that in \cite{Chen2019ICLR}. ${\ddagger}$ note that our AdarGCN achieves an even higher accuracy of \textbf{72.24} with six GCN layers (see Figure \ref{fig_deeper}).}
\label{FSL_comp}
\vspace{-0.1in}
\end{table}

\subsubsection{Comparison to FSL Baselines}

The following FSL baselines are selected: (1) State-of-the-art GCN-based FSL methods \cite{GNN,EGNN,Gidaris_2019_CVPR}; (2) Representative/latest FSL methods (w/o GCN) \cite{ProtoNet,MAML,RelationNet,Ravichandran_2019_ICCV,bertinetto2019meta,IMP,Chen2019ICLR,lee2019meta}. The comparative results under conventional FSL are shown in Table~\ref{FSL_comp}. It can be seen that: (1) Our AdarGCN-FSL method yields 3--5\% improvements over the latest GCN-based FSL methods \cite{GNN,EGNN,Gidaris_2019_CVPR}, validating the effectiveness of adaptive aggregation for GCN-based FSL. (2) The improvements achieved by our method over the state-of-the-art FSL baselines \cite{Ravichandran_2019_ICCV,bertinetto2019meta,IMP,Chen2019ICLR,lee2019meta} range from 1\% to 6\%, showing that AdarGCN has a great potential  for FSL even with sufficient and clean training samples, due to its ability to limit the negative effect of outlying samples.

\vspace{-0.1cm}
\subsubsection{Further Evaluations}

\begin{figure}[t]
\centering
\subfigure[Training]{
\includegraphics[width=0.46\linewidth]{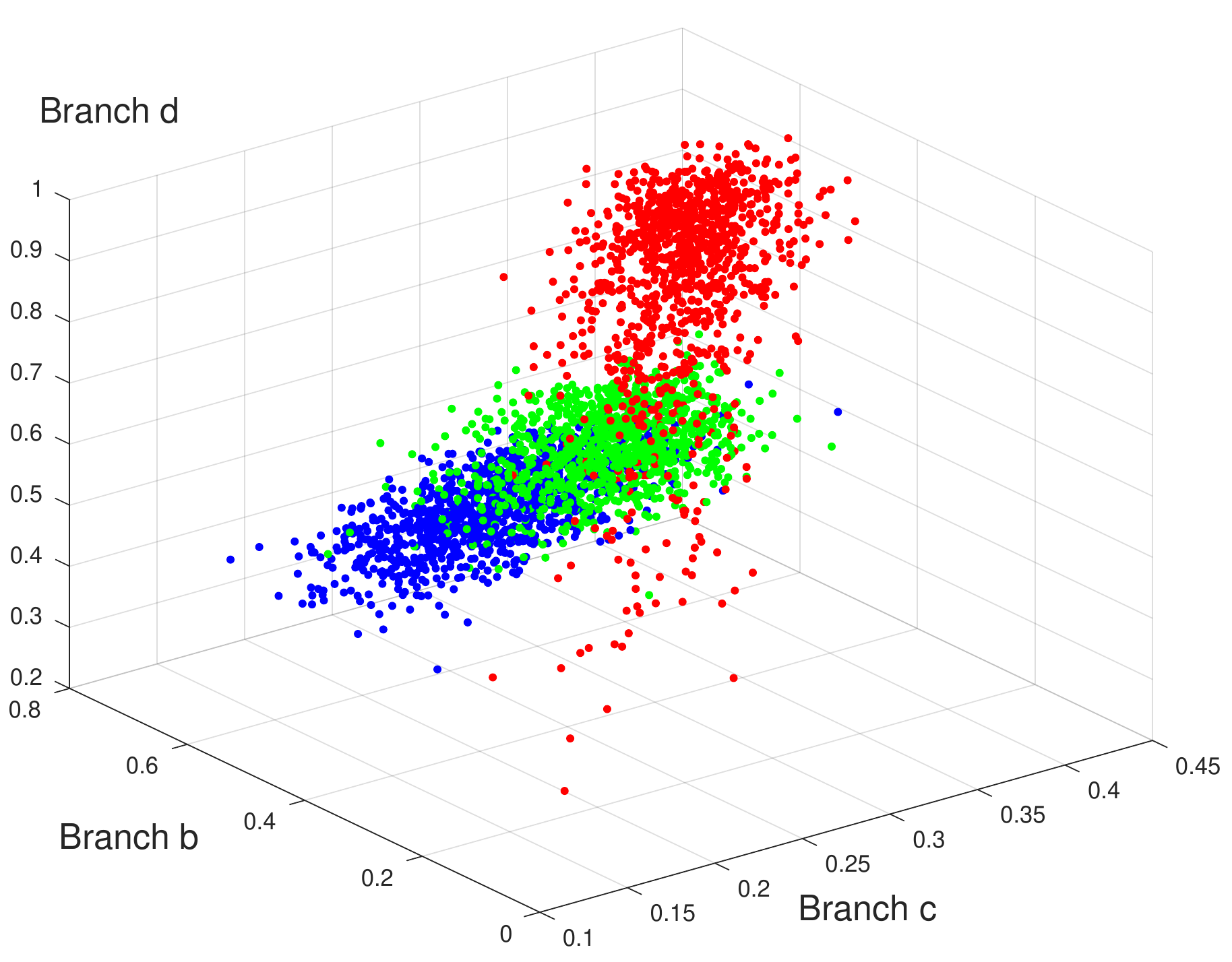}
}%
\subfigure[Test]{
\includegraphics[width=0.46\linewidth]{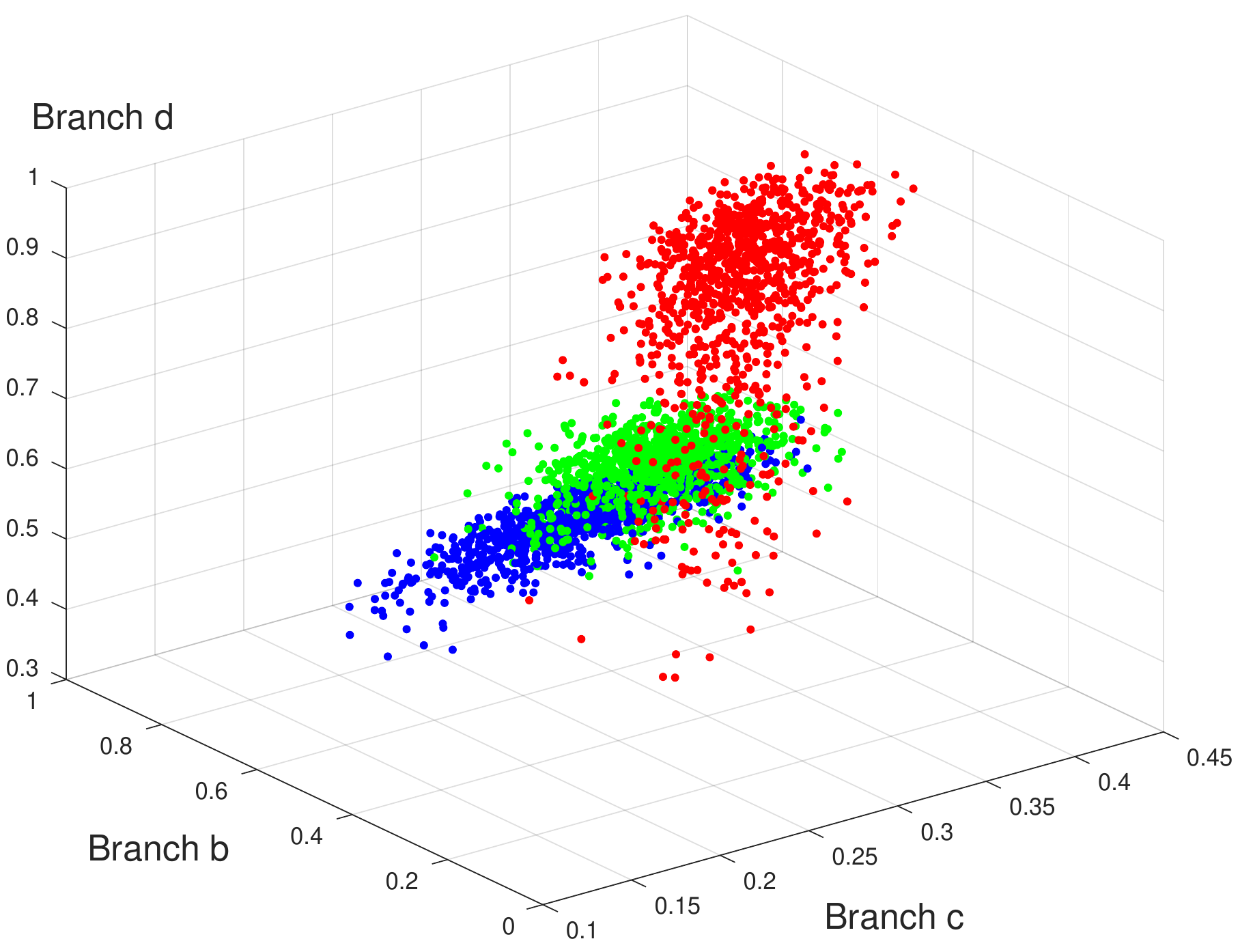}
}
\caption{Illustration of weight distribution on the three branches \textbf{b, c, d} of different GCN layers obtained by our adaptive aggregation module over mini-ImageNet. The \textcolor{red}{red}, \textcolor{green}{green}, and \textcolor{blue}{blue} points denote the weights of GCN layer \textcolor{red}{1}, \textcolor{green}{2}, and \textcolor{blue}{3}, respectively.}
\label{fig_weight}
\vspace{-0.08in}
\end{figure}

\noindent\textbf{Visualization of Adaptive Aggregation}. By randomly sampling 1,000 query images respectively form the training set and the test set, we visualize the weights of the three branches \textbf{b, c, d} of different GCN layers obtained by our adaptive aggregation module (see Figure~\ref{fig:GNN_struct}). The visualization results over mini-ImageNet are presented in Figure~\ref{fig_weight}. It shows that each GCN layer has a significantly different weight distribution. This provides direct evidence that adaptive aggregation is indeed needed in GCN-based FSL. Further, it is also noted that the weight of branch \textbf{c} is forced to be significantly larger than those of the other two branches for the outlying samples so that their negative effect can be effectively limited (see the suppl. material).  

\begin{figure}[t]
\centering
\includegraphics[width=0.85\linewidth]{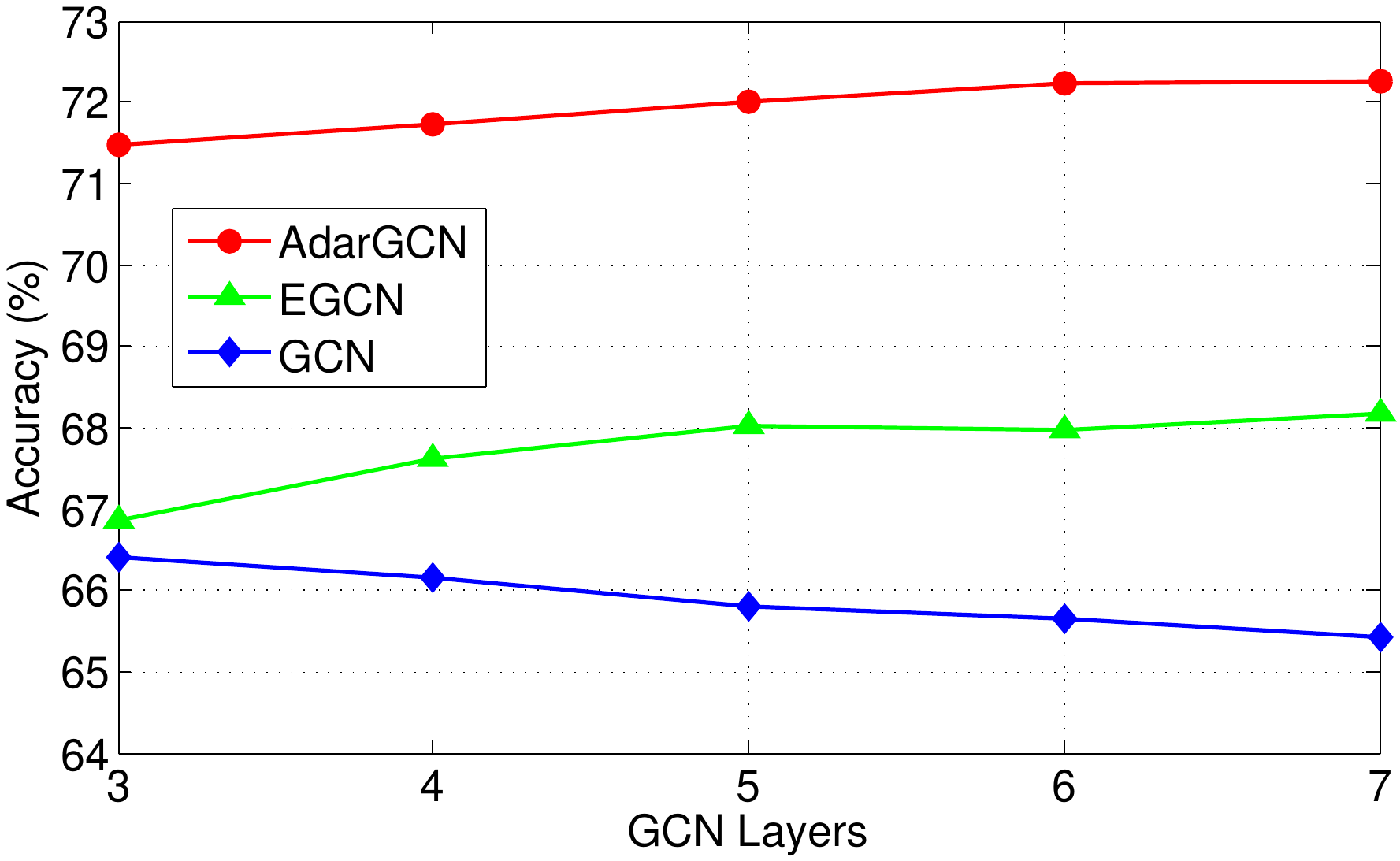}
\vspace{-0.03in}
\caption{Comparative results among the three latest GCN-based FSL methods with deeper GCNs over mini-ImageNet. }
\label{fig_deeper}
\vspace{-0.1in}
\end{figure}

\noindent\textbf{FSL with Deeper GCN}. In all above experiments, each GCN-based FSL method uniformly sets the number of GCN layers to 3, because it is well-known that deeper GCNs often lead to performance degradation. However, since both adaptive aggregation and skip connection are included in our AdarGCN model, it is possible to solve the FSL task with deeper AdarGCN. To explore the challenging problem of FSL with deeper GCNs, we provide the comparative results among the three latest GCN-based FSL methods (i.e. GCN \cite{GNN}, EGCN \cite{EGNN}, and our AdarGCN) in Figure~\ref{fig_deeper}, where the number of GCN layers ranges from 3 to 7. As expected, the performance of GCN \cite{GNN} drops when it goes deeper. However, both EGCN \cite{EGNN} and our AdarGCN achieve performance improvements when more GCN layers are stacked, and our AdarGCN consistently outperforms EGCN. This can be explained as: our AdarGCN leverages both adaptive aggregation and skip connection, while only skip connection is concerned in EGCN. 

\vspace{-0.1cm}
\section{Conclusion}

We have defined a new few-shot few-shot learning (FSFSL) setting. To overcome the training source class data scarcity problem, we chose to augment the training data by crawling sufficient images from the web. Since the crawled images are noisy, we then proposed a GCN-based LDN method to clean the crawled noisy images. Further, with the cleaned web images and the original clean training images as the new training set, we proposed a GCN-based FSL method. For both the LDN and FSL tasks, we designed an AdarGCN model which can perform adaptive aggregation to deal with noisy training data. Extensive experiments demonstrate that our AdarGCN outperforms the state-of-the-art alternatives under both FSL settings.

{
\bibliographystyle{ieee_fullname}
\bibliography{gcn_bib}
}

\end{document}